\documentclass[journal]{IEEEtran}
\usepackage{amsmath,amsfonts}
\usepackage{algorithmic}
\usepackage{array}
\usepackage[caption=false,font=normalsize,labelfont=sf,textfont=sf]{subfig}
\usepackage{textcomp}
\usepackage{stfloats}
\usepackage{url}
\usepackage{verbatim}
\usepackage{graphicx}
\usepackage{svg}
\usepackage{cite}
\usepackage{bbm}
\usepackage{booktabs}
\usepackage{multirow}
\usepackage[backref]{hyperref}
\def\BibTeX{{\rm B\kern-.05em{\sc i\kern-.025em b}\kern-.08em
    T\kern-.1667em\lower.7ex\hbox{E}\kern-.125emX}}
\usepackage{balance}
\usepackage{subfiles}

\begin{document}
\title{Leveraging the Video-level Semantic Consistency of Event for Audio-visual Event Localization}
\author{Yuanyuan Jiang, Jianqin Yin*, Yonghao Dang 
\thanks{Corresponding author: Jianqin Yin}
\thanks{Yuanyuan Jiang, Jianqin Yin, and Yonghao Dang are with the School of Artificial Intelligence, Beijing University of Posts and Telecommunications, Beijing 100876, China (e-mail: jyy@bupt.edu.cn; jqyin@bupt.edu.cn; dyh2018@bupt.edu.cn).}}

\markboth{Journal of \LaTeX\ Class Files,~Vol.~18, No.~9, October~2022}%
{Shell \MakeLowercase{\textit{et al.}}: Leveraging the Video-level Semantic Consistency of Event for Audio-visual Event Localization}

\maketitle

\begin{abstract}
Audio-visual event (AVE) localization has attracted much attention in recent years. Most existing methods are often limited to independently encoding and classifying each video segment separated from the full video (which can be regarded as the segment-level representations of events). However, they ignore the semantic consistency of the event within the same full video (which can be considered as the video-level representations of events). In contrast to existing methods, we propose a novel video-level semantic consistency guidance network for the AVE localization task. Specifically, we propose an event semantic consistency modeling (ESCM) module to explore video-level semantic information for semantic consistency modeling. It consists of two components: a cross-modal event representation extractor (CERE) and an intra-modal semantic consistency enhancer (ISCE). CERE is proposed to obtain the event semantic information at the video level. Furthermore, ISCE takes video-level event semantics as prior knowledge to guide the model to focus on the semantic continuity of an event within each modality. Moreover, we propose a new negative pair filter loss to encourage the network to filter out the irrelevant segment pairs and a new smooth loss to further increase the gap between different categories of events in the weakly-supervised setting. We perform extensive experiments on the public AVE dataset and outperform the state-of-the-art methods in both fully- and weakly-supervised settings, thus verifying the effectiveness of our method. The code is available at \url{https://github.com/Bravo5542/VSCG}.
\end{abstract}


\section{Introduction}
\begin{figure}[!t]
\centering
\includegraphics[width=3.48in]{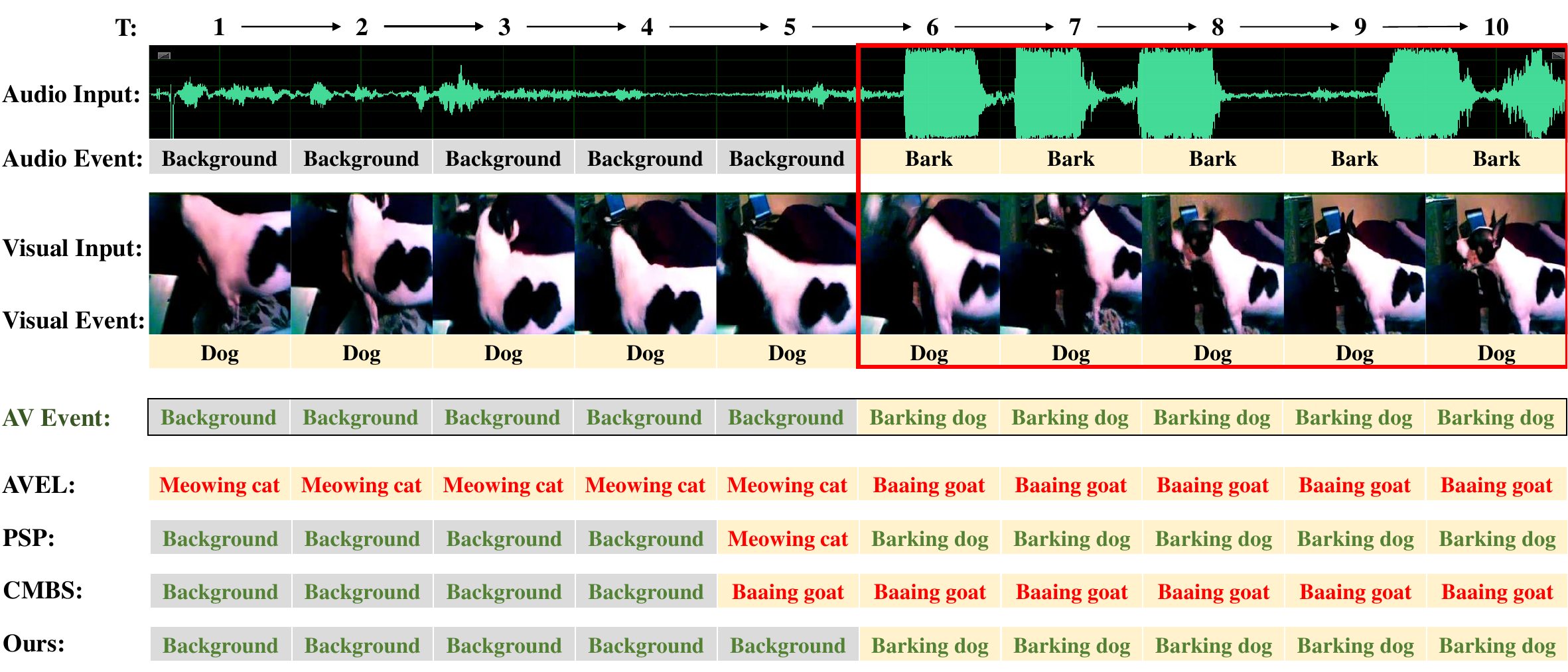}
\caption{Illustration example of the AVE localization task. The red box denotes the segment containing an AVE in which the sounding object is visible and the sound is audible. In the example, only when we see the dog and hear it barking can we localize them (the 5th to 10th segments) as a ``barking dog" AVE, and the remaining are recognized as background. ``AV Event" denotes the ground truth label. The bottom four rows are the predicted results of AVEL\cite{tian2018audio}, PSP\cite{zhou2021positive}, CMBS\cite{xia2022cross}, and our method.}
\label{fig1}
\end{figure}

\IEEEPARstart{R}{esearch} on the audio-visual event (AVE) localization task has proven that joint audio-visual representations can facilitate understanding unconstrained videos \cite{owens2016ambient,stein1989behavioral,tian2018audio,xu2020cross,zhou2021positive,III-3-mahmud2023ave}. Specifically, AVE is defined as an event that is audible and visible simultaneously (as shown in Figure \ref{fig1}). AVE localization requires the model to identify the AVE category and localize its boundary in the temporal dimension. It is essential for understanding, recommending, and searching video content, especially for short video platforms.

A number of approaches have been proposed for AVE localization. Early models \cite{tian2018audio,lin2019dual,lin2020audiovisual,xuan2020cross,ramaswamy2020see,Xue2021CSA,Liu2022DMIN,Xuan2021TIP} mainly focused on mining complementarity between modalities and fusing heterogeneous features. They attempt to align audio and visual information by designing various cross-modal attention modules. However, the performance is inferior since the background class segments cannot be effectively identified, as the AVEL\cite{tian2018audio} method in Figure \ref{fig1} shows. On this basis, some current methods \cite{xu2020cross,zhou2021positive,xia2022cross, III-5-feng2023css} are dedicated to filtering out the unpaired samples (i.e., the background segments) since the audio and visual contents can be temporally inconsistent at the segment level. They adopt a consistent pairs selection or background suppression scheme, which significantly increased the distinction between background segments and segments containing an AVE.

Although promising results have been achieved in previous works, a universal problem remains in previous works: they mainly focus on the segment-level representations of fused audio-visual modalities but overlook the semantic consistency of events across unimodal segments within a short video. As the prediction results of AVEL\cite{tian2018audio} show in Figure \ref{fig1}, separate encoding of segment-level features results in incoherent event semantics throughout the entire video (from meowing cat to baaing goat). Moreover, disregarding video-level information restricts the model's ability to learn more discriminative semantics and the broader context that exists beyond individual segments or modalities. As Figure \ref{fig1} shows, although some segments have representative auditory content (barking), the previous methods were still confused by the visual content. They misidentified the dog as other visually similar categories, e.g., goat or cat. One of the reasons is that existing methods mainly focused on increasing the difference between background and foreground classes while ignoring the need to increase the gap between different foreground classes. However, AVEs often belong to semantically similar or identical categories in the same video. Thus, the video-level representation of the event with discriminative semantics can assist in the category identification of the remaining unimodal segments.

In contrast to previous approaches, we adopt a novel global perspective when addressing AVE localization. As illustrated in Figure \ref{figc}, instead of treating individual segments in isolation, we consider the entire video as a cohesive unit. As is known, there is usually one central theme within a short video, i.e., the semantics are typically consistent throughout that video. We argue that the semantics can be considered a hidden prior or guide for identifying the events, thus providing improvements for AVE localization. Therefore, we propose an event semantic consistency modeling (ESCM) module to explore the video-level semantic information of the AVE for semantic consistency modeling. It consists of two key components.

\begin{figure}[!t]
\centering
\includegraphics[width=3.48in]{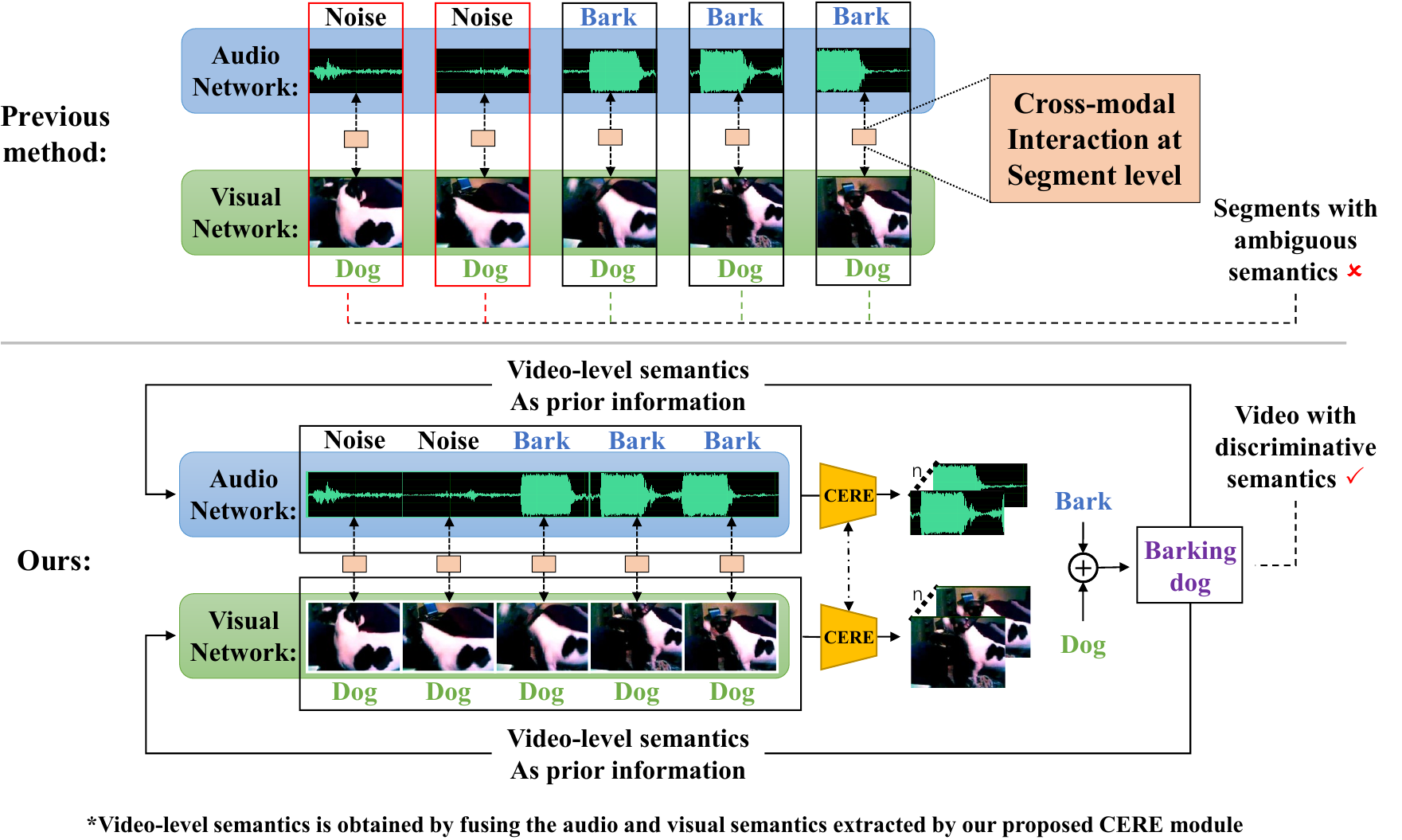}
\caption{Comparison of our video-level semantic guiding approach with previous segment-level encoding approaches. The semantics of the audio-visual content of the first two segments and the visual content of the last three segments are ambiguous, causing the previous method to mistake them for cat or goat. However, the video-level AVE representation has more discriminative semantics, which will help to locate the remaining segments more accurately.}
\label{figc}
\end{figure} 

First, we propose an intra-modal semantic consistency enhancer (ISCE) to enhance semantic consistency within a single modality by leveraging the representation of the video-level AVE, which carries more discriminative semantics than unimodal segments. Event semantic consistency refers to the coherence and consistency of the semantic information related to events across all segments. Hence, we address this by incorporating the video-level AVE representation as a hidden prior for modeling the temporal dynamics of events within each modality. Specifically, ISCE utilizes the video-level AVE representation as a common guide to two separate GRUs in the temporal modeling of each modality. Doing so helps to ensure that the semantic information associated with events remains consistent within each modality, leading to a more coherent and comprehensive representation of events in the overall system.

Second, to obtain the representation of video-level AVE, \textit{i.e.}, video-level semantics, we propose a novel CNN-based cross-modal event representation extractor (CERE), which processes the visual and audio modalities along the temporal dimension. In particular, the video-level audio (visual) event representation is learned by calculating the feature aggregation of all segments to abstract the video audio (visual) semantics. Note that these two CEREs are sharing parameters. In this way, CERE learns representations of temporally synchronized AVEs. Next, we can obtain the video-level AVE representation by fusing the audio event representation and the visual event representation, which carries the integral semantics of the video.

In conclusion, ESCM leverages video-level semantic information to discriminate between different AVEs and further models the dynamic but closely interconnected development of events.

In addition, we propose two losses that enable ESCM to be better applied to both fully and weakly-supervised learning settings. On the one hand, a novel negative pair filter loss is introduced in the fully-supervised setting. The negative pair filter loss enhances the filtering effect on background classes by employing a gating mechanism while increasing the similarity between positive audio-visual pairs. On the other hand, we propose a smooth loss in the weakly-supervised setting. Specifically, we motivate the model to maximize inter-class differences during training by introducing less discriminative class outputs. We integrate all components together and present a video-level semantic consistency guidance network (VSCG), which achieves state-of-the-art performances of 79.7\% and 74.8\% (accuracy) in the fully- and weakly-supervised settings, respectively.

In summary, the contributions of this paper are as follows:

(1) We propose an event semantic consistency modeling (ESCM) module that leverages the video-level AVE representation to guide the semantic continuity modeling of audio and visual features. To achieve this goal, we propose a cross-modal event representation extractor (CERE) to extract effective video-level AVE representations. Furthermore, an intra-modal semantic consistency enhancer (ISCE) is proposed to exploit the video-level AVE representation to maintain a consistent representation of events throughout all segments.

(2) We propose a new negative pair filter loss and smooth loss in fully and weakly-supervised settings, respectively, both of which make the ESCM further improve the distinction of different audio-visual pairs under each setting.

(3) Built upon these modules, we present a video-level semantic consistency guidance (VSCG) network. The experimental results demonstrate that our method outperforms the state-of-the-art methods in both fully- and weakly-supervised tasks on the AVE dataset.

\begin{figure*}[t]
\centering
\includegraphics[width=6.6in]{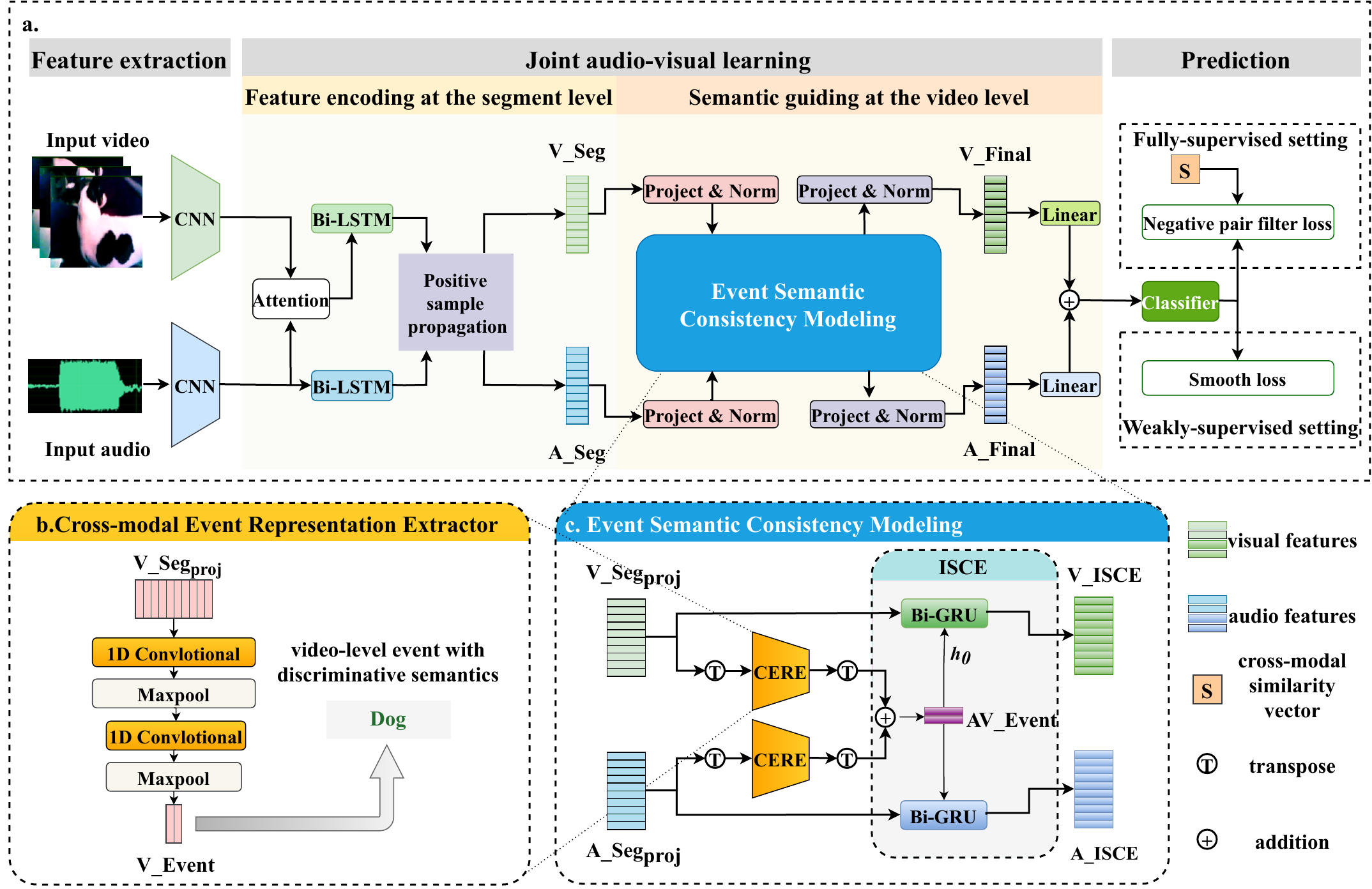}
\caption{The proposed video-level semantic consistency guidance network. (a) The main pipeline of our model. The joint audio-visual learning consists of two parts: the feature encoding at the segment level consisting of audio-guided visual attention, LSTM and the PSP\cite{zhou2021positive} and the semantic guiding at the video level achieved by our proposed event semantic consistency modeling (ESCM) module. (b) Illustration of the cross-modal event representation extractor (CERE) module. We utilize 1D convolutional networks to aggregate all video segments to obtain the video-level event semantic representations. (c) Illustration of the ESCM module consists of CERE and ISCE (intra-modal semantic consistency enhancer), and note that the illustrated CERE modules are shared between audio and visual modalities.}
\label{fig2}
\end{figure*} 

\section{Related Works}
{\bf{Audio-visual Representation Learning:}} In recent years, many works have explored audio-visual representation learning, which can be divided into two main tasks. Some \cite{Owens2018ECCV,lee2021cross,Gan2015CVPR,Zeng2019ICCV,hu2019deep,zellers2022merlot} are dedicated to better video understanding at the video level through the utilization of both visual and auditory information. Zellers \textit{et al}. \cite{zellers2022merlot} constructed a large multimodal pretraining model using masking operations for modal alignment to characterize video information. Many other works have focused on predicting whether video frames and audio are temporally aligned and the fusion of video and audio information at the segment level. Hu \textit{et al}. \cite{hu2019deep} proposed a deep multimodal clustering (DMC) network to perform elaborate correspondence learning among audio and visual components. Existing methods often prioritize visual content for classification, overlooking the significance of auditory information. To learn discriminative features for a classifier, it is essential to judge whether the audio and visual signals depict the same event. Some recent methods attempt to evaluate the correspondences by measuring the audio-visual similarity \cite{hu2019deep,arandjelovic2017look,arandjelovic2018objects,cheng2020look,Wu2022WSavDetec} while ignoring that as an information carrier, each video, especially for short videos, always has a specific theme, i.e., the video-level event semantics.

{\bf{Sounding Object Localization:}} Sounding object localization aims to localize visual regions that are relevant to the given sound. By digging into the correspondence between deep audio and visual features, recent methods have provided promising solutions to this problem \cite{arandjelovic2018objects,afouras2020self,tian2021cyclic,valverde2021there,Zhao2018ECCV,Gan2019ICCV,III-2-lin2023unsupervised}. Afouras \textit{et al}. \cite{afouras2020self} utilized a contrastive loss to train the model in a self-supervised learning way. Qian \textit{et al}. \cite{qian2020multiple} achieved the same goal by using the class activation map (CAM) derived from a weakly-supervised approach. Song \textit{et al}. \cite{song2022self}, inspired by 3D object detection under self-supervision, proposed a contrastive learning strategy based on data augmentation after matching only positive sample pairs to achieve localization of sounding objects in the visual region with self-supervised learning.

Most AVE localization methods draw on the sounding object localization task of visually localizing vocal objects to facilitate the recognition of events. In this paper, we utilized the semantic consistency of events to guide the learning of visual and audio modalities, which can ensure that the model better focuses and localizes sounding objects.

{\bf{Audio-visual Event Localization:}} With the development of joint audio-visual learning, the AVE localization task was formally proposed by Tian \textit{et al}. \cite{tian2018audio} with the first public dataset AVE. They proposed an audio-guided visual attention module (AGVA) and a dual multimodal residual network to fuse information over two modalities. Xu \textit{et al}. \cite{xu2020cross} proposed an audio-guided spatial-channel attention (CMRA) module that can guide the model to focus on event-relevant visual regions. Zhou \textit{et al}. \cite{zhou2021positive} proposed a positive sample propagation (PSP) module that constructs an all-pair similarity map between each audio and visual segment and removes the entries that are below a preset similarity threshold. Similar to \cite{zhou2021positive}, Fan \textit{et al}. \cite{III-5-feng2023css} proposed a consistent segment selection approach to enhance segment-level semantic consistency across modalities. However, they ignored the event semantic consistency within modalities. Xia \textit{et al}. \cite{xia2022cross} proposed a background suppression scheme based on CMRA \cite{xu2020cross}, whose network is more complex than the others. Similar to \cite{xu2020cross,xia2022cross}, Liu \textit{et al}. \cite{Liu2022DMIN} improves the temporal-spatial-channel attention module and further establishes the cross-modal dense relationship. Recently, Geng \textit{et al}. \cite{III-4-geng2023dense} introduced a new dense AVE localization task, making a valuable contribution to the field. Nonetheless, it fundamentally differs from AVE localization and is, therefore, beyond the scope of this paper.

Although the above works achieve superior performance on the AVE localization task, they ignore the holistic and continuous occurrence of events in the natural environment.
While impressive works \cite{wang2020dual,xiao2021boundary,wang2021structured} excel in the video temporal grounding (VTG) task by leveraging segment-level features to improve unimodal video frame localization, it is important to note that VTG uses explicit textual semantic information. In contrast, AVE localization relies on hidden semantics within multimedia content with audio-visual disparities and involves processing heterogeneous features from two modalities.
Unlike those works, we propose a video-level semantic consistency guidance network by further exploring the semantic consistency of AVEs at the video level specifically tailored for audio-visual event localization.

\section{Methods}
\subsection{Problem Setting}
AVE localization requires locating the segment where the audio-visual content is synchronized and then identifying the specific event category within that segment. Specifically, the given video \textit{S} is divided into \textit{T} non-overlapping consecutive segments $ {\{\boldsymbol{S}_t^v,\boldsymbol{S}_t^a\}}^{T}_{(t=1)} $ (\textit{t} represents the temporal segment index), where each segment is set to 1 second long. AVE localization takes visual and audio features $ \boldsymbol{S}_t^v\in \mathbb{R}^{d_a} $,$ \boldsymbol{S}_t^a\in \mathbb{R}^{H\times W\times d_v} $ extracted by pretrained models as inputs. The model is required to predict the event label of each segment as $\{\boldsymbol{y}_t={y_t^c|y_t^c\in \{0,1\},\sum\nolimits_{c=1}^{C}y_t^c=1\}}\in \mathbb{R}^C$. Here, C is the total number of AVE events (including the background). In the fully-supervised setting, the label of each segment is available during the training phase, and the label of the entire video is represented as $ \boldsymbol{Y}^{fully}=[\boldsymbol{y}_1;\boldsymbol{y}_2;\ldots;\boldsymbol{y}_T]\in \mathbb{R}^{T\times C} $.

In the weakly-supervised setting, we only have access to a video-level tag $\boldsymbol{Y}^{weakly}\in \mathbb{R}^{1\times C}$ that represents which event category is contained in the video. Despite this limitation, we are required to predict segment-level labels during inference, which poses a greater challenge.

\subsection{Overall Pipeline}
The architecture of the proposed method is illustrated in Figure \ref{fig2}.a, including feature extraction, joint audio-visual learning, and prediction. Joint audio-visual learning is the crucial process of our proposed model, which consists of two steps: feature encoding at the segment level and semantic guiding at the video level. The second step will be achieved through the event semantic consistency modeling (ESCM) module. The feature encoding at the segment level includes the initial feature fusing and background screening. The proposed ESCM module exploits video-level semantics as prior information to identify interesting events and empirically models the development of events in the temporal dimension. Lastly, the updated audio-visual features modeled by ESCM are fused and then sent to the final classification module, predicting which video segments contain an event and the event category.

\subsection{Feature Encoding at the Segment Level}\label{PSP}
Since the elements contained in the video and audio of each segment do not correspond exactly one-to-one, e.g., the silent object and out-of-picture sound, we need to encode the features at the segment level for initial audio-visual fusion. As seen in Figure \ref{fig2}.a, we adopt an attention module and positive sample propagation strategy as our segment-level feature encoding structure. First, in the encoding module, an audio-guided visual attention (AGVA) module \cite{tian2018audio} is used to perform an early fusion of audio-visual information, making the model focus on the visual regions with a higher correlation with the corresponding audio segments. The initial encoding of the temporal relationships is then completed by using a single-layer Bi-LSTM \cite{schuster1997bidirectional}. Next, we utilize the positive sample propagation module (PSP) \cite{zhou2021positive} to enhance the encoded features by diminishing audio and visual samples that have weak connections. In this way, the audio and visual features are aggregated with the positively correlated segments to obtain the segment-level encoded features $ \boldsymbol{A}^{seg}\in \mathbb{R}^{T\times d_s} $, $\boldsymbol{V}^{seg}\in \mathbb{R}^{T\times d_s}$.

Next, the obtained $\boldsymbol{A}^{seg}\in \mathbb{R}^{T\times d_s} $ and $\boldsymbol{V}^{seg}\in \mathbb{R}^{T\times d_s} $ are further projected and normalized. Each project and norm block consists of a linear projection layer followed by ReLU, dropout with a rate of $ r_s $, and a LayerNorm layer.

\subsection{Event Semantic Consistency Modeling}\label{ESCM}
As typical natural signals, both audio and visual features have strong redundancy and noise, which are useless and even distracting for segments of the other modality. However, continuous video often has the same semantics, especially for short videos. The semantics can be regarded as a hidden prior for identifying interesting events. Therefore, the key solution to this problem lies in enabling the model to focus on the critical features of the video-level multimodal event that carry more discriminative semantics than the unimodal segment.

As shown in Figure \ref{fig2}.c, the ESCM module considers the semantic continuity of the event and leverages the video-level semantics to model the semantic consistency of segment-encoded features. First, the video-level audio and visual event representations are extracted by the cross-modal event representation extractor. Then, the audio event representation and visual event representation are fused to obtain the video-level AVE representation, which carries the integral video-level semantics of the event. Subsequently, the intra-modal semantic consistency enhancer utilizes the video-level AVE representation as prior information to guide the temporal modeling within each modality. Thus, the semantic consistency of events is enhanced by modeling the semantic continuity of unimodal events.

\subsubsection{Cross-modal Event Representation Extractor (CERE)}\label{CERE}
Most existing approaches encode each segment to obtain segment-level predictions while ignoring the semantic consistency of events at the video level. In contrast to previous methods, we propose CERE to obtain video-level event representations that carry more discriminative semantics, i.e., video-level semantics, than random unimodal segments.

The CERE consists of two consecutive CNN \cite{lecun1998gradient} blocks in the audio and visual branch each. As shown in Figure \ref{fig2}.b, here, we only display the visual branch for simplicity. We first transpose the segment-level encoded visual feature $ \boldsymbol{V}^{seg}\in \mathbb{R}^{T\times d_s} $, and each column $ {{\boldsymbol{v}^{seg}}^\top}\in \mathbb{R}^{d_s\times 1} $ denotes one segment. CERE takes the transposed visual feature as input, and then the features can be subsequently operated in the time dimension instead of the feature dimension. Manipulating in the temporal dimension allows for a larger receptive field in time when extracting event representations, leading to a better understanding of video-level events. After the CERE module, we obtain a video-level representation of the visual event (dominated by the salient features) that can abstract the visual semantics of the entire video. Specifically, for one CNN block, we perform convolution at $ {\boldsymbol{V}^{seg}}^\top ({\boldsymbol{A}^{seg}}^\top) $ in the time dimension with the 1D convolutional network, followed by maximum pooling to capture salient features of the event. After two CNN blocks, we obtain the video-level visual (audio) event $ \boldsymbol{E}^{v}(\boldsymbol{E}^{a}) $ that carries discriminative semantic information. In particular, to obtain the cross-modal synchronized event representation, we use two identical CEREs to extract the video-level audio event representation and visual event representation. In this way, the modality with more representative video-level semantics can directly influence the representation learning of another modality, and the convolution in the temporal dimension ensures the synchronization of audio and visual events. The shared CERE module allows the synchronous mapping of the input audio and visual features to a unified feature space to ensure effective fusion of features from different modalities. After CERE, we obtain the video-level audio event representation $ \boldsymbol{E}^{a}\in \mathbb{R}^{d_e\times 2} $ and the video-level visual event representation $ \boldsymbol{E}^{v}\in \mathbb{R}^{d_e\times 2} $, computed by:
\begin{equation}
\label{eq1}
\boldsymbol{E}^{a}=CNN_{Block2} (CNN_{Block1} ({\boldsymbol{A}^{seg}}^\top)).
\end{equation}
\begin{equation}
\label{eq2}
\boldsymbol{E}^{v}=CNN_{Block2} (CNN_{Block1} ({\boldsymbol{V}^{seg}}^\top)).
\end{equation}
\begin{equation}
\label{eq3}
CNN_{Block\textit{i}} = MAX(\delta(K\ast f)).
\end{equation}

\noindent where $ f $ refers to $ {\boldsymbol{A}^{seg}}^\top\in \mathbb{R}^{d_s\times T} $ or $ {\boldsymbol{V}^{seg}}^\top\in \mathbb{R}^{d_s\times T} $ in our model, $ K $ is the learning kernel parameters with a size of $ T/2 $ (the size of the output channel is $ d_e$), $ \ast $ is the convolutional operator, $ \delta $ is the ReLU activation function, and $ MAX $ denotes the max pooling operation along the temporal dimension. Note that audio and visual branches share the same CNN blocks with the same learning parameters, and we believe that this helps to learn audio-visual semantic representations belonging to the cross-modal synchronized event rather than separate single modal events that are significant at different time steps, which ensures that the learned $ \boldsymbol{E}^{a} $ and $ \boldsymbol{E}^{v} $ are semantically complementary.

\subsubsection{Intra-modal Semantic Consistency Enhancer (ISCE)}

ISCE leverages video-level semantics to enhance semantic consistency across all segments within each modality. Since we are concerned with events that are both audible and visible in the video, a representation of video-level events that contain cross-modal semantics can aid in the learning of the intra-modal segment-level features. Recurrent neural network (RNN) is designed for modeling sequential data to capture temporal dependencies within audio or visual features. Specifically, ISCE utilizes context-related global representations to initialize RNN, which could further enhance the consistency and rationality of the unimodal event's dynamic but closely interconnected development.

In the ISCE module, we utilize two separate GRUs \cite{cho2014learning} to take $\boldsymbol{A}^{seg}$ and $\boldsymbol{V}^{seg}$ as inputs to model the semantic consistency in each modality. For an input feature vector ${\boldsymbol{f}_t}$ at time step $t$, the GRU updates a hidden state vector $\boldsymbol{h}_t$ and output vector $\boldsymbol{y}_t$ as follows:
 \begin{equation}
\label{eq4}
\boldsymbol{z}_t=\sigma(\boldsymbol{W}_z\cdot [\boldsymbol{h}_{t-1},\boldsymbol{f}_t]).
\end{equation}
\begin{equation}
    \label{eq5}
    \boldsymbol{r}_t=\sigma(\boldsymbol{W}_r\cdot [\boldsymbol{h}_{t-1},\boldsymbol{f}_t]).
\end{equation}
\begin{equation}
    \label{eq6}
    \tilde{\boldsymbol{h}}_t = \tanh{(\boldsymbol{W}\cdot[\boldsymbol{r}_t\odot\boldsymbol{h}_{t-1},\boldsymbol{f}_t])}.
\end{equation}
\begin{equation}
    \label{eq7}
    \boldsymbol{h}_t=\boldsymbol{z}_t\odot\boldsymbol{h}_{t-1}+(1-\boldsymbol{z}_t)\odot \tilde{\boldsymbol{h}}_t.
\end{equation}
\begin{equation}
    \label{eq8}
    \boldsymbol{y}_t=\sigma(\boldsymbol{W}'\boldsymbol{h}_t)
\end{equation}

Most previous approaches usually initialize the hidden state vector $\boldsymbol{h}_0$ of the GRU \cite{cho2014learning} to zeros to indicate the beginning of the sequence, leading to the model treating the initial features of each sequence without discrimination so that the salient features of the event are suppressed at the beginning to some extent. However, for the AVE task, events of the same type follow a similar development pattern over time, i.e., semantic continuity. For example, when we are observing the ``helicopter coming towards us" (the video-level event), the proportion of the helicopter in the picture changes from small to large, and the sound we hear also changes from weak to loud, no matter what type of helicopter it is. Therefore, the video-level semantic information of the event can guide visual and audio segment-level feature learning in the temporal dimension.

\textbf{Complementary cross-modal information.} Events in the natural environment often possess audio-visual complementarity. In Figure \ref{fig2}.c, we describe how to leverage video-level event representations in this section. Thanks to the parameter sharing between two CEREs in Sec. \ref{CERE}, we can fuse the extracted $\boldsymbol{E}^{a}$ and $\boldsymbol{E}^{v}$ to obtain the video-level AVE representation that carries the integral semantics of the multimedia video. As demonstrated in Figure \ref{fig2}.c, we fuse $\boldsymbol{E}^{a}$ and $\boldsymbol{E}^{v}$ to obtain the video-level AVE representation $ \boldsymbol{E}^{av}\in \mathbb{R}^{2\times d_e} $ that contains semantic information from both modalities, formulated as follows:
 \begin{equation}
\label{eq9}
\boldsymbol{E}^{av}=\frac{1}{2}\times({\boldsymbol{E}^{v}}^\top+{\boldsymbol{E}^{a}}^\top).
\end{equation}

\textbf{Introducing a priori information in temporal modeling.} Previous approaches have modeled video sequence features in a semantically agnostic manner, but with the extracted video-level semantic information, we can add it as prior information to the modeling process of the video sequences. Based on the mentioned observation, we propose the ISCE module to leverage the video-level AVE representation $\boldsymbol{E}^{av}$ as prior knowledge which carries discriminative semantics. Specifically, we take $\boldsymbol{E}^{av}$ as the common initial hidden state of two independent GRUs to guide the modeling of the dynamics of audio and visual events. In this way, the GRU can focus on significant features that are semantically consistent with the video-level event instead of treating each feature indiscriminately at the beginning. Furthermore, $\boldsymbol{E}^{av}$ includes both visual and audio semantics of the event. Hence, the audio (visual) features can be used as supplementary information to assist the model in learning more robust semantically continuous representations when modeling the temporal changes of visual (audio) features. The final features are formulated as follows:

\begin{equation}
\label{eq10}
\boldsymbol{A}^{ISCE}=\mathbf{GRU}(\boldsymbol{A}^{seg},\boldsymbol{E}^{av}).
\end{equation}
\begin{equation}
\label{eq11}
\boldsymbol{V}^{ISCE}=\mathbf{GRU}(\boldsymbol{V}^{seg},\boldsymbol{E}^{av}).
\end{equation}
where $ \boldsymbol{E}^{av}\in \mathbb{R}^{2\times d_e} $, $ \boldsymbol{A}^{seg}, \boldsymbol{V}^{seg}\in \mathbb{R}^{T\times d_s} $. Referring to Eq. (\ref{eq4})-(\ref{eq8}), $\boldsymbol{h}_0=\boldsymbol{E}^{av}, \boldsymbol{f}_t\in\{\boldsymbol{a}^{seg}_t,\boldsymbol{v}^{seg}_t\}, \boldsymbol{y}_t\in\{\boldsymbol{a}^{ISCE}_t,\boldsymbol{v}^{ISCE}_t\}$. $\boldsymbol{A}^{ISCE}$ and $\boldsymbol{V}^{ISCE}\in \mathbb{R}^{T\times d_i} $ are the final representations that contain both discriminative semantic characteristics at the video level and clear temporal boundaries at the segment level. Our experiments revealed that employing more complex techniques, such as LSTMs, did not significantly improve model performance. Thus, to balance efficiency and accuracy, a single-layer bidirectional GRU is used, and $d_i$ equals $2d_e$.
 
Similarly, we feed the obtained $\boldsymbol{A}^{ISCE}$ and $\boldsymbol{V}^{ISCE}$ into the project and norm blocks, respectively. Then, late fusion is performed, which can be formulated as follows:
\begin{equation}
\label{eq12}
\boldsymbol{F}^m=\mathcal{D}[\delta(\boldsymbol{M}^{ISCE}\boldsymbol{W}_1^m)],\boldsymbol{m}\in \{\boldsymbol{a},\boldsymbol{v}\}.
\end{equation}
\begin{equation}
\label{eq13}
\boldsymbol{F}^{v \leftrightarrow a}=\frac{1}{2}[\mathcal{N}(\boldsymbol{F}^a)+\mathcal{N}(\boldsymbol{F}^v)].
\end{equation}
where $ \boldsymbol{W}_1^v, \boldsymbol{W}_1^a\in \mathbb{R}^{d_i\times d_f} $ denotes the learnable parameters in the linear layers, $\delta$ is the ReLU activation function, $ \mathcal{D} $ is the dropout with a rate of $r_i$, and $ \mathcal{N}(\cdot) $ represents layer normalization. The output $\boldsymbol{F}^{v \leftrightarrow a}\in \mathbb{R}^{T\times d_f} $ can be more distinguished for the following classification module since ISCE focuses on the discriminative semantics of the event at the video level and better models its development pattern.

\subsection{Classification and Objective Function}
\textbf{Negative pair filter loss under fully-supervised.} Following \cite{xu2020cross,xia2022cross}, we decompose the supervised AVE localization task into two subtasks. First, one predicts whether the segment contains AVEs, i.e., whether the segment belongs to the background category, according to the event-relevant score $ \boldsymbol{o}_t\in \mathbb{R}^{T} $. Then, the other predicts the category of events according to event-category label $ \boldsymbol{o}_c\in \mathbb{R}^{C-1} $. Specifically, $\boldsymbol{o}_t$ and $\boldsymbol{o}_c$ can be calculated as:
\begin{equation}
\label{eq14}
\boldsymbol{o}_t=\mathcal{S}(\boldsymbol{F}^{v\leftrightarrow a}\boldsymbol{W}_3).
\end{equation} 
\begin{equation}
\label{eq15}
\boldsymbol{o}_c= \tilde{\boldsymbol{f}}^{v\leftrightarrow a}\boldsymbol{W}_4.
\end{equation}
where $ \boldsymbol{W}_3\in \mathbb{R}^{d_f\times 1} $, $ \boldsymbol{W}_4\in \mathbb{R}^{d_f\times (C-1)} $ are the learnable parameters, $\mathcal{S}$ represents the squeeze operation, and $ \tilde{\boldsymbol{f}}^{v\leftrightarrow a}\in \mathbb{R}^d_f $ is the max pooling result from $ \boldsymbol{F}^{v\leftrightarrow a} $. Because the total $C$ categories of events include one background class, we determine whether the $t-th$ video segment contains an event, i.e., belongs to the background, according to the value of $\boldsymbol{o}_t$. Thus, we only need to predict the AVE category number as $C-1$ instead of $C$.
\begin{equation}
\label{eq16}
\boldsymbol{s}=\frac{\boldsymbol{V}^{ISCE}\odot \boldsymbol{A}^{ISCE}}{{\|\boldsymbol{V}^{ISCE}\odot \boldsymbol{A}^{ISCE}\|}_1}. 
\end{equation}

In addition, inspired by \cite{zhou2021positive}, we compute the $ {\ell}_{1} $ normalized similarity vector $\boldsymbol{s}\in \mathbb{R}^{T}$ between the visual features $\boldsymbol{V}^{ISCE}$ and audio features $\boldsymbol{A}^{ISCE}$ of each segment, as shown in Eq. (\ref{eq16}), where $\odot$ denotes element-wise multiplication. It is optimized by audio-visual pair similarity loss $\mathcal{L}_{avps}$ proposed in \cite{zhou2021positive}, which encourages the ESCM module to maintain the cross-modal segment-level relevance when temporal modeling. During training, we obtain both the segment-level event category $\boldsymbol{Y}_{tc}\in \mathcal{R}^{T\times(C-1)}$ provided by annotations and the background label $\boldsymbol{y}_t=\{y_t |y_t\in \{0,1\},t=1,2,\ldots,T\}\in \mathbb{R}^T,$ where $y_t$ represents whether the $t-th$ segment is an AVE or background. Moreover, we adopt $\boldsymbol{y}_{tl}\in \mathbb{R}^T$, which equals the ${\ell}_1$ normalized $\boldsymbol{y}_t$. Thus, the overall objective function negative pair filter loss includes category loss and background loss, which is:
\begin{equation}
\label{eq17}
\mathcal{L}_{fully}=\overbrace{-\frac{1}{T(C-1)}\sum\nolimits_{t=1}^T\sum\nolimits_{c=1}^{C-1}\boldsymbol{Y}_{tc}\log (\boldsymbol{O}_{tc})}^{\mathcal{L}_c}+\mathcal{L}_b.
\end{equation}
\begin{equation}
\label{eq18}
\mathcal{L}_b=\overbrace{\mathcal{L}_{BCE}(\boldsymbol{o}_t,\boldsymbol{y}_t)}^{\mathcal{L}_t}+\overbrace{\mathcal{L}_{MSE}(\boldsymbol{s},\boldsymbol{y}_{tl})}^{\mathcal{L}_{avps}}.
\end{equation}
where $\mathcal{L}_c$ is the cross-entropy loss of the network output $\boldsymbol{O}_{tc}$ and segment-level label $\boldsymbol{Y}_{tc}$, and $\mathcal{L}_t$ refers to the binary cross-entropy loss of the network output $\boldsymbol{o}_t$ and background label $\boldsymbol{y}_t$. $\mathcal{L}_{avps}$ computes the mean squares error between the $\ell_1$ normalized similarity vector $\mathcal{\boldsymbol{s}}$ and $\ell_1$ normalized background label $\boldsymbol{y}_{tl}$ so that it promotes the network to produce similar features for a pair of audio and visual components if the pair contains an AVE. Because $\mathcal{L}_b$ consists of $\mathcal{L}_t$ and $\mathcal{L}_{avps}$, which encourages the model to increase the gap between positive pairs (containing an AVE) and negative pairs (background), the negative pair filter loss $\mathcal{L}_{fully}$ allows better exploitation of the segment-level encoding to optimize the entire network. During the inference phase, the background segment is presumably featured by smaller event-relevant scores, so we simply adopt a thresholding method to filter out the background. Specifically, if $o_t>\tau_b$, then the $t-th$ video segment is predicted as the $o_c$ foreground classes. Otherwise, the $t-th$ video segment is classified as background. $\tau_b$ is set to 0.7.

\textbf{Smooth loss under weakly-supervised.} 
Following \cite{zhou2021positive}, we adopt a weighting branch on the weakly-supervised classification module that enables the model to highlight the differences between synchronized audio-visual pairs belonging to different event categories. The output $\boldsymbol{O}_c^w\in \mathbb{R}^{1\times C}$ represents the AVEs contained in the video.

The objective function adopts a smooth loss, which is implemented by binary cross-entropy (BCE) loss, formulated as:
\begin{equation}
\label{eq19}
\mathcal{L}_{weak}=\lambda\mathcal{L}_{bce}+\mathcal{L}_{s-{bce}}.
\end{equation}
\begin{equation}
\label{eq20}
\mathcal{L}_{bce}=\mathcal{L}_{BCE} (\boldsymbol{O}_c^w,\boldsymbol{Y}_c).
\end{equation}
\begin{equation}
\label{eq21}
\mathcal{L}_{s-{bce}}=\mathcal{L}_{BCE}(s(\boldsymbol{O}_c^w),\boldsymbol{Y}_c).
\end{equation}
where $\lambda$ is a hyperparameter to balance the two losses. $s$ is a softmax function, and $ \boldsymbol{Y}_c\in \mathbb{R}^{C\times \mathbbm{1}} $ denotes the video-level event-category label. We use the softmax function twice to generate a smoother probability distribution $ s(\boldsymbol{O}_c^w) $ over $C$ event categories, which promotes the model to capture more discriminative features in the weakly-supervised setting. In particular, by introducing less discriminative category outputs as a basis for judgment in the inference phase, the model is thus motivated to further increase the gap between different categories of events in the training phase.

\begin{table}
\begin{center}
\caption{Comparisons with state-of-the-art methods in two supervised settings, measured by accuracy(\%) on the AVE dataset. * indicates that the number was obtained with the same attention module as ours.}
\label{tab1}
\begin{tabular}{ccc}
\toprule
Method & Fully-supervised & Weakly-supervised\\
\midrule
AVEL \cite{tian2018audio} & 72.7 & 66.7\\
AVSDN \cite{lin2019dual} & 72.6 & 67.3\\
CMAN \cite{xuan2020cross} & 73.3 & 70.4\\
DAM \cite{wu2019dual} & 74.5 & $-$\\
AVRB \cite{ramaswamy2020see} & 74.8 & 68.9\\
AVIN \cite{ramaswamy2020makes} & 75.2 & 69.4\\
CSA \cite{Xue2021CSA} & 76.5 & 70.2\\
AVT \cite{lin2020audiovisual} & 76.8 & 70.2\\
CMRA \cite{xu2020cross} & 77.4 & 72.9\\
PSP \cite{zhou2021positive} & 77.8 & 73.5\\
DCA \cite{Xuan2021TIP} & 79.0 & $-$\\
CSS-Net \cite{III-5-feng2023css} & 73.5* & 74.2*\\
CMBS \cite{xia2022cross} & 79.3 & 74.2\\
DMIN \cite{Liu2022DMIN} & 79.6 & 74.3\\
\midrule
\textbf{VSCG} (Ours) & \textbf{79.7} & \textbf{74.8}\\
\bottomrule
\end{tabular}
\end{center}
\end{table}
 
\section{Experiments}
\subsection{Experimental Setup}
\textbf{Dataset:} The AVE dataset originating from the AudioSet \cite{gemmeke2017audio} contains 4,143 videos covering various real-life scenes. It can be divided into 28 event categories, e.g., church bell, male speech, acoustic guitar, and dog barking. Each video sample is evenly partitioned into 10 segments lasting one second each. The AVE boundary on the segment level and the event category are provided.

\textbf{Evaluation Metrics:} The category label of each segment is required to be predicted in both fully- and weakly-supervised settings. Following \cite{tian2018audio}, we adopt the classification accuracy of the predicted segment-level event category as the evaluation metric.

\textbf{Implementation Details:} We use VGG-19 \cite{simonyan2014very} pretrained on ImageNet \cite{russakovsky2015imagenet} to extract the visual features of size $7\times7\times512$. Specifically, 16 frames are sampled from each one-second segment. We extract the visual feature maps from each frame and use the average map as the visual feature for this segment. For audio features, we first process the raw audio into log-mel spectrograms and then extract the 128-dimensional audio features using a VGG-like network \cite{hershey2017cnn} pretrained on AudioSet \cite{gemmeke2017audio}. In addition, rate $r_s$ in Sec. \ref{PSP} is empirically set to $0.2$, and rate $r_i$ in Sec. \ref{ESCM} is empirically set to $0.2$ and $0.5$ in the fully- and weakly-supervised setting, respectively. Weight $\lambda$ in Eq. (\ref{eq19}) is empirically set to $2$. We use a batch size of $128$, and the optimizer is Adam. Our model is trained on an NVIDIA GeForce GTX 1080 and implemented in PyTorch.

\begin{table}
\begin{center}
\caption{Comprehensive comparisons with current state-of-the-art methods in fully-supervised settings. The top-2 results are highlighted.}
\label{tab4}
\begin{tabular}{ccc}
\toprule
Method & Accuracy(\%) & Trainable Params.(M)\\
\midrule
AVT \cite{lin2020audiovisual} & 76.8 & 15.8\\
CMRA \cite{xu2020cross} & 77.4 & 15.9\\
PSP \cite{zhou2021positive} & 77.8 & \textbf{1.7}\\
CMBS \cite{xia2022cross} & 79.3 & 14.4 \\
DMIN \cite{Liu2022DMIN} & \underline{79.6} & 17.7\\
\midrule
\textbf{VSCG} (Ours) & \textbf{79.7} & \underline{2.6}\\
\bottomrule
\end{tabular}
\end{center}
\end{table}

\subsection{Comparisons with State-of-the-art Methods}
We challenge our method against current fully- and weakly-supervised SOTA methods on AVE tasks. For a fair comparison, we use the same settings as those in previous methods.

We compare our method with the state-of-the-art methods in Table \ref{tab1}, where our method achieves superior results: the classification accuracies are 79.7\% and 74.5\% for the fully- and weakly-supervised settings, respectively. Notably, our method significantly surpasses the baseline model PSP-Net \cite{zhou2021positive} by a large margin (1.9\% and 1.0\%, respectively), demonstrating the effectiveness of our proposed method. In comparison with CCS-Net \cite{III-5-feng2023css}, whose main contribution lies in consistent segment selection, our method outperforms it by 6.2\% in the fully-supervised setting and 0.6\% in the weakly-supervised setting when the same attention module AGVA \cite{tian2018audio} is used. These results strongly validate the superiority of our video-level semantic consistency modeling approach over the segment-level approach. Even though the current state-of-the-art method DMIN \cite{Liu2022DMIN} has a sophisticated cross-modal attention module, our method still outperforms them by 0.1\% and 0.5\% in fully- and weakly-supervised settings, respectively. As shown in Table \ref{tab4}, the number of trainable parameters in our model is less than one-sixth of the current SOTA (from 2.6 M to 17.7 M). See Supplemental Material I. \textit{Supplemental Experiments} for more comprehensive comparisons of memory and time needed during training with SOTAs. When compared to the baseline model\cite{zhou2021positive}, we achieved a significant improvement (1.9\%) at the cost of a small increase in the number of parameters. This can be attributed to the proposed ESCM module, which, although simple, takes full advantage of the semantic consistency of events in the entire video.

\begin{table}
\begin{center}
\caption{Ablation studies of different modules on AVE dataset. The top-2 results are highlighted.}
\label{tab2}
\begin{tabular}{ccc}
\toprule
Method & Fully-supervised & Weakly-supervised\\
\midrule
w/o ESCM & 77.9 & 72.9\\
w/o ISCE & 76.4 & 71.9\\
ISCE w/ \it{Avg} & 77.8 & 73.0\\
ISCE w/ \it{Max} & \underline{78.2} & \underline{73.4}\\
ISCE w/ non-shared CERE & 77.2 & 72.8\\
ISCE w/ CERE (Ours) & \textbf{79.7} & \textbf{74.8}\\
\bottomrule
\end{tabular}
\end{center}
\end{table}

\begin{table}
\begin{center}
\caption{Ablation studies of the proposed ESCM module. * indicates the number is reproduced by us.}
\label{tab5}
\begin{tabular}{ccc}
\toprule
Method & Fully-supervised & Weakly-supervised\\
\midrule
AVEL \cite{tian2018audio} & 72.7* & 66.7*\\
AVEL w/ PSP \cite{zhou2021positive} & 76.6* & 72.9*\\
AVEL w/ ESCM & \textbf{77.5} & \textbf{72.9}\\
\midrule
PSP-Net \cite{zhou2021positive} & 76.6* & 72.9*\\
PSP-Net \cite{zhou2021positive} w/ ESCM & \textbf{78.8} & \textbf{73.8}\\
\midrule
CMBS \cite{xia2022cross} & 79.2* & 73.3*\\
CMBS \cite{xia2022cross} w/ ESCM & \textbf{79.6} & \textbf{73.4}\\
\bottomrule
\end{tabular}
\end{center}
\end{table}

\subsection{Ablation Studies}
To verify the effectiveness of the proposed method, we remove the individual module from the network and re-evaluate it in both supervised settings, as shown in Table \ref{tab2}.

\textbf{The effectiveness of the ESCM.} To evaluate the effect of ESCM, we remove the ESCM module and denote it as ``w/o ESCM". We observe from the table that the performance drops in both settings significantly. Specifically, compared to the full model, the accuracy of the ``w/o ESCM" decreases by 1.8\% (from 79.7\% to 77.9\%) and 1.9\% (from 74.8\% to 72.9\%) for each setting, which distinctly validates the effectiveness of the ESCM.

To further validate the effectiveness of the proposed ESCM module, a comparison experiment was conducted against 3 baseline models \cite{tian2018audio}\cite{zhou2021positive}\cite{xia2022cross}. Note that the experimental settings are identical to the baseline model accordingly, and we simply add our ESCM module to the very end of each baseline model. As shown in Table \ref{tab5}, our ESCM module (denoted as ``AVEL w/ ESCM") surpasses PSP module \cite{zhou2021positive} (denoted as ``AVEL w/ PSP") by 0.9\% in the fully-supervised setting and achieves the same accuracy (72.9\%) in the weakly-supervised setting. Moreover, our proposed ESCM module can bring further improvements to the recent SOTAs PSP-Net\cite{zhou2021positive} and CMBS\cite{xia2022cross}, as seen in Rows 4-7. The results prove two points. First, our proposed ESCM module is superior to the baseline PSP module (from 76.6\% to 77.5\%). Second, our proposed ESCM module can bring further significant accuracy improvements upon AVEL\cite{tian2018audio} (4.8\% and 6.2\%, respectively), PSP-Net\cite{zhou2021positive} (2.2\% and 0.9\%, respectively), and CMBS\cite{xia2022cross} (0.4\% and 0.1\%, respectively). We notice that the improvement upon CMBS is limited. The reasons may be that 1) CMBS diverges notably from the common pipeline, particularly under weakly-supervised conditions, necessitating some design adjustments to fully unleash the potential of ESCM. 2) Some issues in the dataset also affect the upper bound of the model performance, see II.B. \textit{Issues of Temporal Boundaries} in the Supplemental Material. These results suggest that our proposed plug-and-play module can benefit multiple AVE localization approaches, as none of them take into account video-level semantics.

\textbf{The effectiveness of the ISCE.} As shown in Table \ref{tab2}, we validate the effectiveness of the ISCE module, which utilizes video-level AVE representation to enhance the semantic consistency of segment sequences. Thus, we remove the utilization of video-level AVE representation, i.e., use zero-initialized GRUs to model the temporal dynamics of each modality, denoted as ``w/o ISCE".  The accuracy of ``w/o ISCE" drops by 3.3\% in the fully-supervised setting and 2.9\% in the weakly-supervised setting compared to our model.

Table \ref{tab2} shows that utilizing video-level semantics, e.g., ``ISCE w/ $\it{Avg}$" and ``ISCE w/ $\it{Max}$", is higher than that of models without it (``w/o ISCE") by a large margin. In addition, ``ISCE w/ $\it{Avg}$", ``ISCE w/ $\it{Max}$" and our ``ISCE w/ CERE" all achieved higher or competitive accuracy than ``w/o ESCM". The lowest accuracies of 76.4\% and 71.9\% are achieved by ``w/o ISCE", which are even 1.5\% and 1.0\% lower than removing the entire ESCM module (denoted as ``w/o ESCM"). The result indicates that simply deepening the temporal modeling of video features is not only useless but also hurts the performance of the model. Experimental results demonstrate that GRUs initialized with event representations related to the input sequence features tend to have better performance.

These results demonstrate the effectiveness of our proposed ISCE module, which leverages video-level semantics as prior knowledge to guide the network to enhance the semantic consistency within each modality.

\textbf{The necessity of the CERE.} Analogously, we validate the effectiveness of CERE by substitution and deformation. The overall model's performance decreases after the change.

For the CERE module, the core lies in the representation of video-level events, i.e., video-level semantics. Thus, we initialize the ISCE in four ways corresponding to four ways of obtaining video-level event representations $\boldsymbol{E}^{av}$.

Specifically, we attempt to obtain $\boldsymbol{E}^{av}$ by performing average pooling (denoted as ``ISCE w/ $\it{Avg}$") and max pooling (denoted as ``ISCE w/ $\it{Max}$") on segment-level encoded video features $\{\boldsymbol{A}^{seg}, \boldsymbol{V}^{seg}\}$. ``ISCE w/ $\it{Max}$" achieved the second-best accuracy of 78.2\% and 73.4\% after our full model (``ISCE w/ CERE"), surpassing ``w/o ESCM" by 0.3\% and 0.5\%, respectively. ``ISCE w/ Max" is similar to our idea of capturing salient features as video-level AVE representations, which tend to have more discriminative semantics than unimodal segments. In particular, the video-level event representation extracted by our proposed parameter-shared CERE module (``ISCE w/ CERE") achieves the highest accuracy of 79.7\% and 74.8\% compared to ``ISCE w/ $\it{Avg}$" and ``ISCE w/ $\it{Max}$", respectively. These results validate the superiority and effectiveness of our proposed CERE module for video-level event representation extraction.

In addition, we use the common CNN blocks in the CERE module for the visual and audio branches, i.e., sharing parameters. To validate the shared CERE, we compare ours with the CERE of the visual and audio branches using different CNN blocks (denoted as ``ISCE w/ non-shared CERE"), which increases the number of parameters in the model. As shown in Table \ref{tab2}, the performance after CERE deformation exceeds ``w/o ISCE" but surprisingly decreases by 2.5\% and 2.0\% compared to ours in fully- and weakly-supervised settings, respectively. This suggests that CERE may be influenced by background information in unimodality. Our proposed parameter-shared CERE effectively avoids capturing background representations, as it considers information from both modalities simultaneously.

These results validate that our proposed parameter-shared CERE ensures the cross-modal complementarity and synchronization of the extracted video-level event representations from the visual and audio modalities.

\begin{table}
\begin{center}
\caption{Method comparison on the AVE dataset in two settings. We evaluate 1) the negative pair filter loss in the fully-supervised setting and 2) the smooth loss in the weakly-supervised setting. The two improvements are implemented on top of our system and PSP-Net. $\ast$ indicates the number is reproduced by us. We use bold font to show the higher performance brought by our technique.}
\label{tab3}
\begin{tabular}{cccc}
\toprule
Setting &Method & VSCG (Ours)& PSP \cite{zhou2021positive}\\
\midrule
\multirow{3}{*}{Fully} 
& $\mathcal{L}_c+\mathcal{L}_t$ & {78.1} & {76.3$\ast$}\\
& $\mathcal{L}_{ce}+\lambda\mathcal{L}_{avps}$ \cite{zhou2021positive} & 78.8 & 76.6$\ast$\\
& $\mathcal{L}_c+\mathcal{L}_t+\mathcal{L}_{avps}$ (Ours) & \textbf{79.7} & \textbf{77.5$\ast$}\\
\midrule
\multirow{2}{*}{Weakly}
& $\mathcal{L}_{bce}$ & {73.8} & {72.9$\ast$}\\
& $2\mathcal{L}_{bce}+\mathcal{L}_{s-bce}$ (Ours) & \textbf{74.8} & \textbf{73.0$\ast$}\\
\bottomrule
\end{tabular}
\end{center}
\end{table}

\textbf{Benefit of the negative pair filter loss $\mathcal{L}_{fully}$ and the smooth loss $\mathcal{L}_{weak}$.} In the fully-supervised setting, we adopt $\mathcal{L}_c+\mathcal{L}_t$, $\mathcal{L}_{ce}+\lambda\mathcal{L}_{avps}$ \cite{zhou2021positive} and $\mathcal{L}_{fully}=\mathcal{L}_c+\mathcal{L}_t+ \mathcal{L}_{avps}$ as the objective function to train the model, respectively. Here, $\mathcal{L}_{ce}$ denotes the cross-entropy between the output (the predicted 29 categories, including background) of the model and the label provided by the dataset. $\lambda$ is a balanced factor that is set to 100 according to \cite{zhou2021positive}.

To verify the effectiveness of the proposed losses, as presented in Table \ref{tab3}, we use our VSCG network and the currently popular PSP network as baselines. In the fully-supervised setting, the results show that the negative pair filter loss $\mathcal{L}_{fully}$ improves the accuracy compared to the previous two loss functions. Particularly compared with the loss in \cite{zhou2021positive}, the improvement brought by $\mathcal{L}_{fully}$ is 0.9\% for both VSCG and PSP. These results validate the effectiveness of the negative pair filter loss $\mathcal{L}_{fully}$,  where $\mathcal{L}_t$ and $\mathcal{L}_{avps}$ work together as an auxiliary restriction to better distinguish between background segments and segments containing AVEs. Specifically, $\mathcal{L}_{avps}$ increases the similarity between audio and visual modalities of segments containing AVEs, while $\mathcal{L}_t$ decreases the probability score when a segment contains no AVEs.

For the weakly-supervised setting, we respectively adopt $\mathcal{L}_{bce}$ and the smooth loss $\mathcal{L}_{weak}=2\mathcal{L}_{bce}+\mathcal{L}_{s-bce}$ as the objective function for model training. As shown in Table \ref{tab3}, the smooth loss significantly improved the accuracy of VSCG by 1.0\%, while the improvement of PSP was faint. This implies a better exploitation of the smoothing loss effect due to the presence of the ESCM module, which obtained more discriminative video features.

\subsection{Qualitative Analysis}
\begin{figure}[!t]
\centering
\includegraphics[width=3.48in]{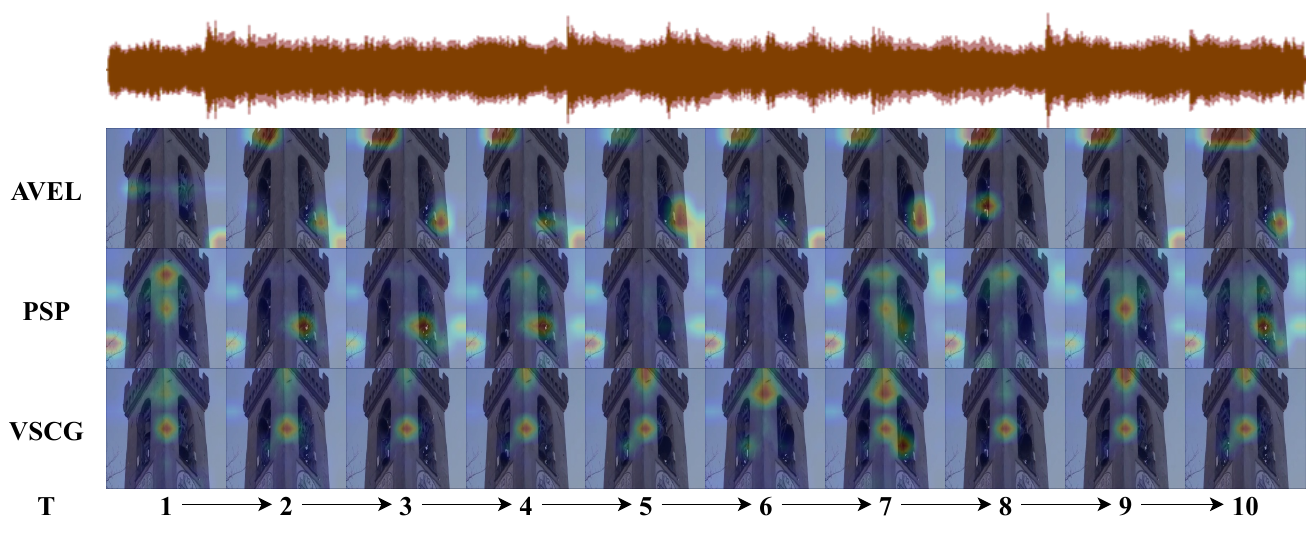}
\caption{A qualitative example of AVE localization in a visually obscured scene. For this video, all ten segments contain the visual and audio signals of the ``ringing church bell" event. We visualize the visual attention in the image stream. It is clear that our method produces more accurate localization and that our attended regions better overlap with the sound sources. We choose intermediate image frames for visualization as an abstract representation of the segment.}
\label{fig3}
\end{figure}

We show the visualization results of the attention module since AVEL \cite{tian2018audio}, PSP \cite{zhou2021positive}, and the proposed VSCG adopt the same auditory-guided visual attention module (AGVA) \cite{tian2018audio}. The results show that the proposed VSCG is better able to focus on the visual areas that are closely related to the sound source. As shown in Figure \ref{fig3}, the sample event itself is elementary, but the presence of occlusion makes it difficult for the machine to utilize the visual information. For the ``church bell ringing" event, VSCG can always focus on the area within the visual area that is closely related to the ``bell ringing", e.g., the area where the bell that emits the ringing is located. In contrast, even though the video content and the auditory content presented in the video are continuously stable, the area of interest of PSP \cite{zhou2021positive} is scattered and changes over time. The AGVA module in \cite{zhou2021positive} even completely focuses on the independent area in the 5th and 6th segments. For AVEL \cite{tian2018audio}, the sound is incorrectly located in the ``sky" area of the background all the time due to the influence of the occlusion.

\begin{figure}[!t]
\centering
\includegraphics[width=3.48in]{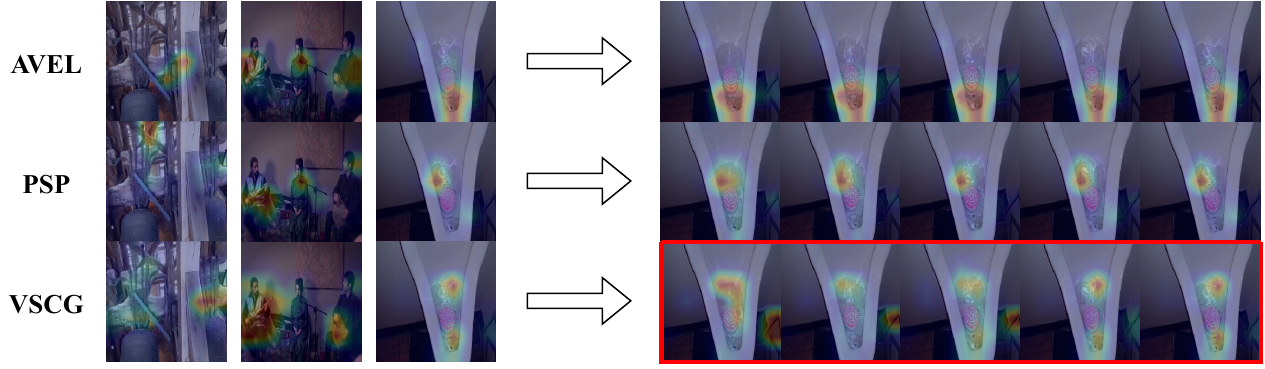}
\caption{A qualitative example of AVE localization in a multi-source scene. For the visualization results on the left, the first column shows two large bells on the left and right sides of the scene; the second column shows that there are three people in the image, but the woman in the middle does not make any sound; in the third column, the water hits the upper wall of the sink and then naturally flows down to hit the lower wall, so there are two sounding points.}
\label{fig5}
\end{figure}

In the example of ``church bell ringing" shown in Figure \ref{fig3}, even with the introduction of auditory modality, it is still challenging to locate the obscured sound source (bell) in the visual area. However, in the proposed VSCG network, the ESCM module enables the model to capture the visual representation of the ``church bell ringing" event as the dependency structure of the ``bell tower and bell". This verifies that our approach learns deep semantic information about the event rather than just the features of modalities. Thus, the actual location of the sounding object can be better localized based on the audio information by our method.

As shown in Figure \ref{fig5}, the proposed method is able to locate more accurately when there are multiple sound-emitting regions in the image. The ``flush" event in the lower panel of Figure \ref{fig5}, due to the vertical structure of the sink, causes the water flowing out to collide with the upper wall first and then downward to impact the lower wall to make a sound. We argue that due to ISCE introducing video-level semantic information as prior knowledge for temporal modeling, the model can more accurately locate the sound-emitting region and learn the changes over time, thereby behaving more correctly to physical laws. As shown in Figure \ref{fig5}, AVEL \cite{tian2018audio} and PSP \cite{zhou2021positive} do not have such a property.

\section{Conclusion and Future Work}
\textbf{Conclusion.} For the AVE localization task, we propose a video-level semantic consistency guidance network, which further models the continuity and plausibility of events as they develop in the time dimension. Compared to previous approaches that focus on analyzing segments independently, we acknowledge the importance of contextual information and the temporal coherence of events within a short video. Our proposed ESCM module consists of two components, CERE and ISCE, for the extraction of video-level AVE representations and their exploitation, respectively. This allows for a more comprehensive understanding of the semantic consistency between different modalities and their contributions to the overall event semantics. Furthermore, for the AVE localization task, we propose the negative pair filter loss and the smooth loss in the fully and weakly-supervised setting, respectively. Extensive experiments validate the effectiveness of our proposed method, and ablation studies validate the effectiveness of the proposed modules.

\textbf{Limitations.} Although our proposed semantic consistency-based approach can significantly improve recognition accuracy, it will misidentify events as semantically similar categories when the semantics of the video in both the audio and visual modalities are highly similar to the wrong category. In the future, we would like to explore video-level event semantic guidance in combination with the text modality, e.g., labels, which contain explicit semantics. Moreover, we will extend video-level event semantic guidance to other multimodal tasks, e.g., audio-visual representation learning and sounding object localization.

\section*{Acknowledgement}
This work was supported partly by the National Natural Science Foundation of China (Grant No. 62173045, 62273054), partly by the Fundamental Research Funds for the Central Universities (Grant No. 2020XD-A04-3), and the Natural Science Foundation of Hainan Province (Grant No. 622RC675).

\bibliographystyle{IEEEtran}
\bibliography{IEEEabrv,references}{}

\begin{IEEEbiography}[{\includegraphics[width=1in,height=1.25in,clip,keepaspectratio]{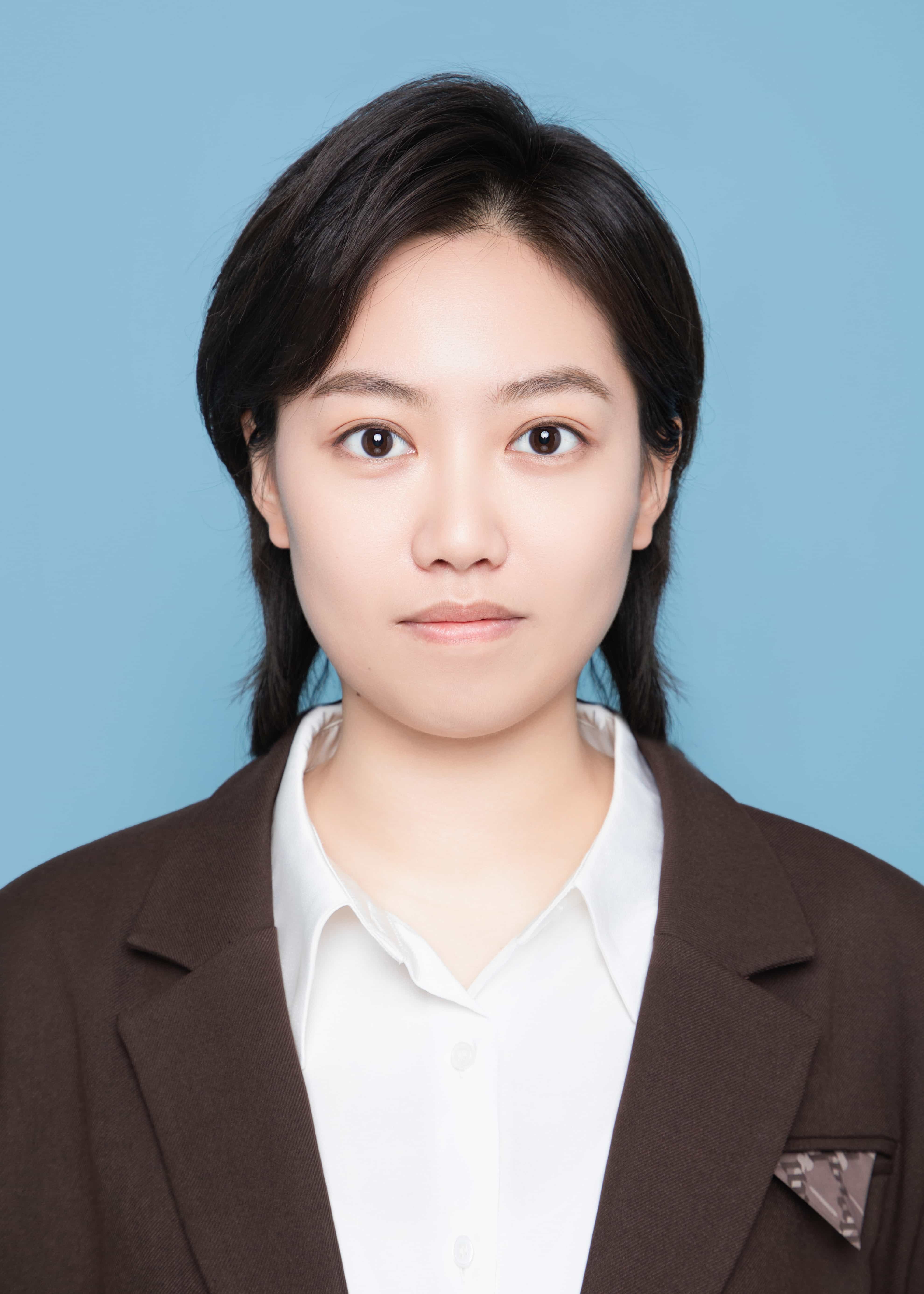}}]{Yuanyuan Jiang}
Yuanyuan Jiang received a B.Eng. degree from Beijing University of Posts
and Telecommunications, Beijing, China, in 2018. She is currently an M.S. candidate with the School of Artificial Intelligence, Beijing University of Posts and Telecommunications, Beijing, China. Her research interests include multimodal learning, computer vision, machine learning, and deep learning. Email: jyy@bupt.edu.cn.
\end{IEEEbiography}
\begin{IEEEbiography}[{\includegraphics[width=1in,height=1.25in,clip,keepaspectratio]{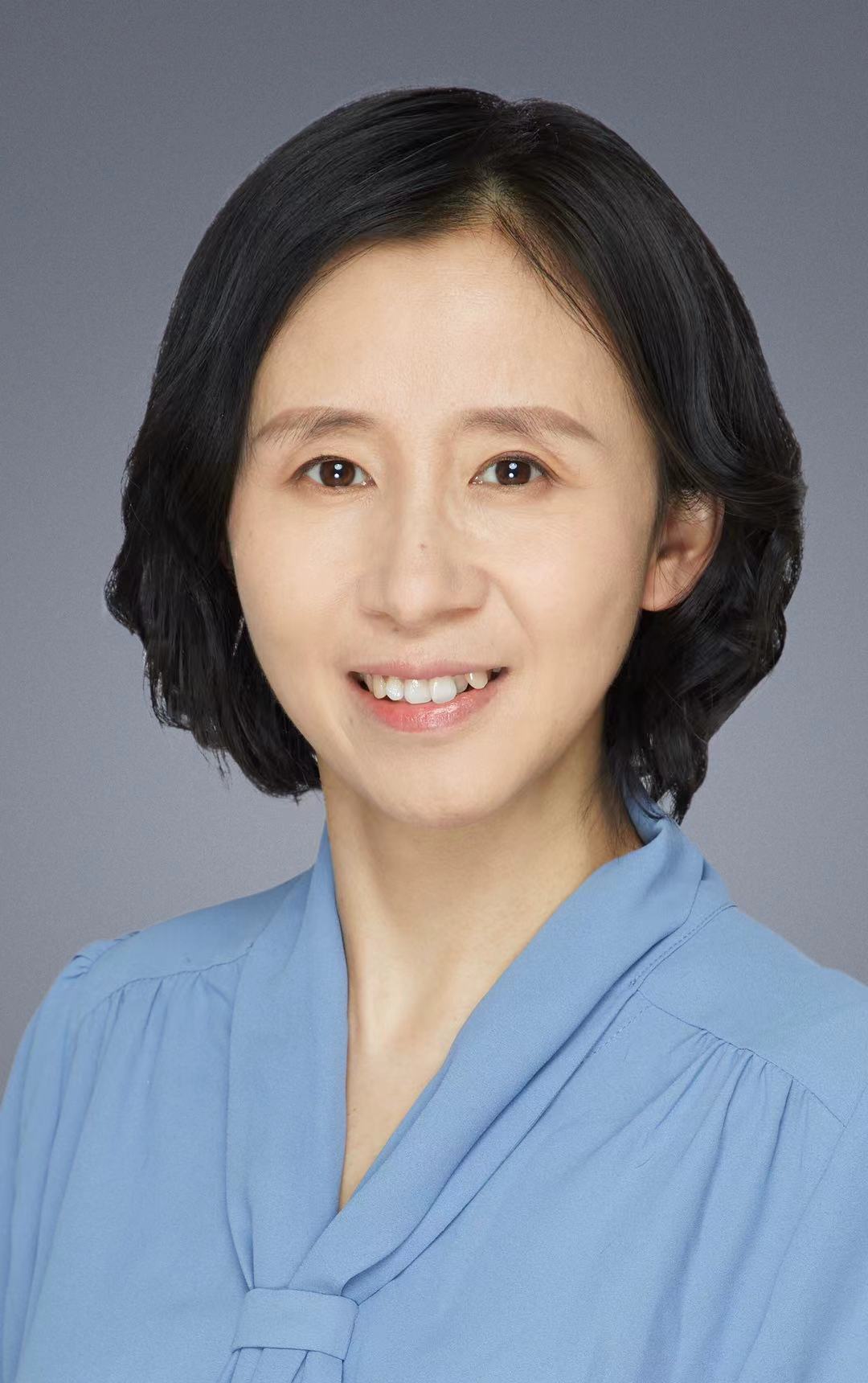}}]{Jianqin Yin}
Jianqin Yin received a Ph.D. degree from Shandong University, Jinan, China, in 2013. She is currently a professor at the School of Artificial Intelligence, Beijing University of Posts and Telecommunications, Beijing, China. Her research interests include service robots, pattern recognition, machine learning, and image processing. Email: jqyin@bupt.edu.cn.
\end{IEEEbiography}
\vspace{-80mm} 
\begin{IEEEbiography}[{\includegraphics[width=1in,height=1.25in,clip,keepaspectratio]{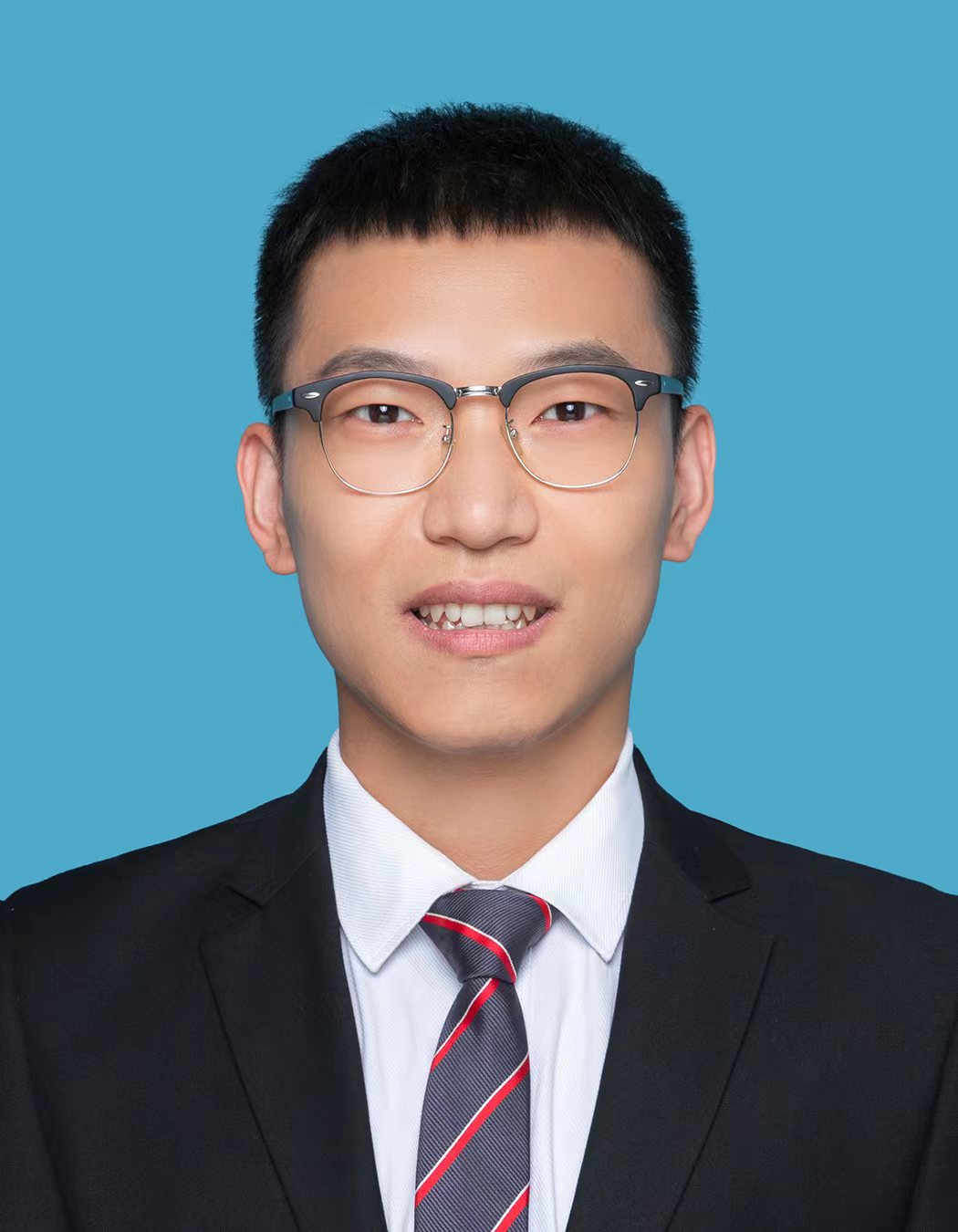}}]{Yonghao Dang}
Yonghao Dang received a B.Eng. degree in computer science and technology from the University of Jinan, Jinan, China, in 2018. He is currently a Ph.D. candidate with the School of Artificial Intelligence, Beijing University of Posts and Telecommunications, Beijing, China. His research interests include computer vision, machine learning, image processing, and deep learning. Email: dyh2018@bupt.edu.cn.
\end{IEEEbiography}


\newpage
\setcounter{section}{0}\section{Supplemental Experiments}
\subsection{Comprehensive comparisons with current SOTAs}

\begin{table}
\begin{center}
\renewcommand{\thetable}{VI}
\caption{Comprehensive comparisons with current state-of-the-art methods in fully-supervised settings. The top-2 results are highlighted. * indicates the number is reproduced by us.}
\label{tabadd}
\resizebox{\linewidth}{!}{
\begin{tabular}{cccc}
\toprule
Method & Accuracy(\%) & Training Mem.(M) &Training Time \\
\midrule
CMBS  & \underline{79.2*} & \underline{3039} & \underline{2h31min}\\
DMIN  & 78.5* & 8079 & 7h20min \\
\midrule
\textbf{VSCG} (Ours) & \textbf{79.7} &\textbf{2623} & \textbf{54min}\\
\bottomrule
\end{tabular}}
\end{center}
\end{table}

We reproduced the recent SOTAs and conducted the comparison of memory and time required during training with ours. Note that the experimental settings, such as the batch size and epoch, used in each article are slightly different. Thus, we conducted experiments not only under the experimental settings used in the corresponding article but also under the experimental settings used in most of the methods as ours. For each method, we adopted the best performance we obtained under both settings as the final result. The experimental results can be seen in Table \ref{tabadd}. Experimental results show that our method achieves 0.5\% better results than the second-best method CMBS in a much shorter training time (from 2h31min to 54min) with a small memory cost (from 3039 MiB to 2623 MiB). We believe that the experimental results further validate the effectiveness and efficiency of our method.

\section{Supplemental Qualitative Analysis}
\subsection{Distracting Visual Content}

\begin{figure}
\centering
\includegraphics[width=3.48in]{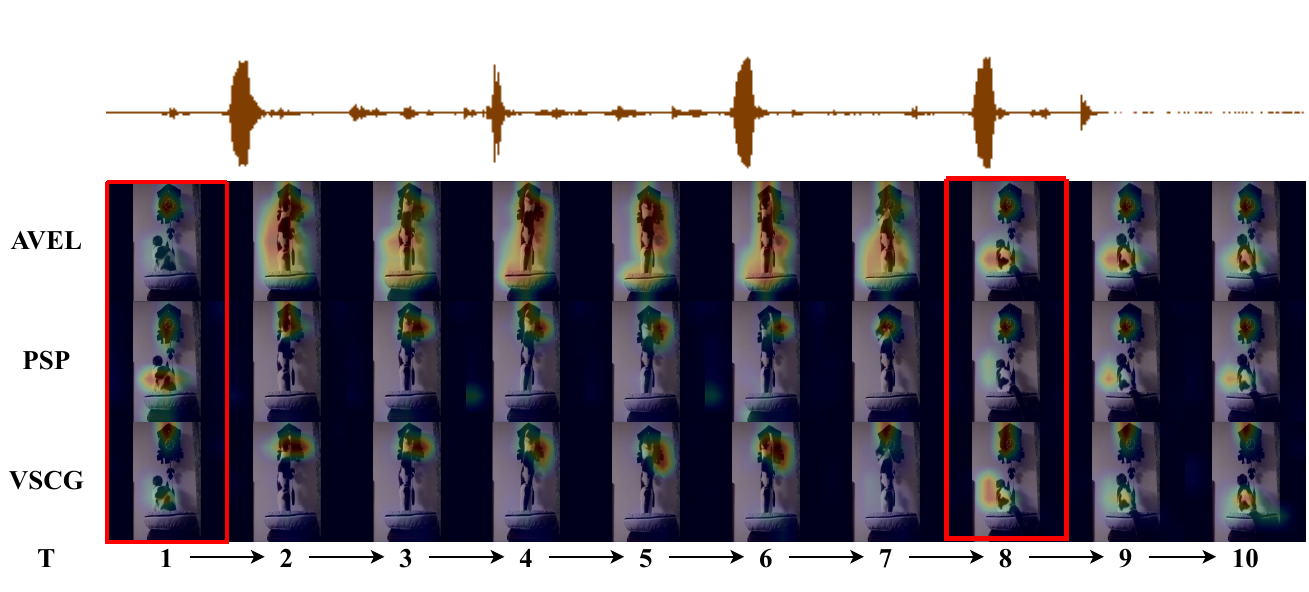}
\renewcommand{\thefigure}{6}
\caption{A qualitative example of AVE localization in a scene where the visual subject is prominent. For this video, all ten segments contain the visual and audio signals of the “striking clock” event. However, the secondary subject, “cat”, is notable but silent in most segments. We choose intermediate image frames for visualization as an abstract representation of the segment.}
\label{fig4}
\end{figure}

\begin{figure}
\centering
\includegraphics[width=3in]{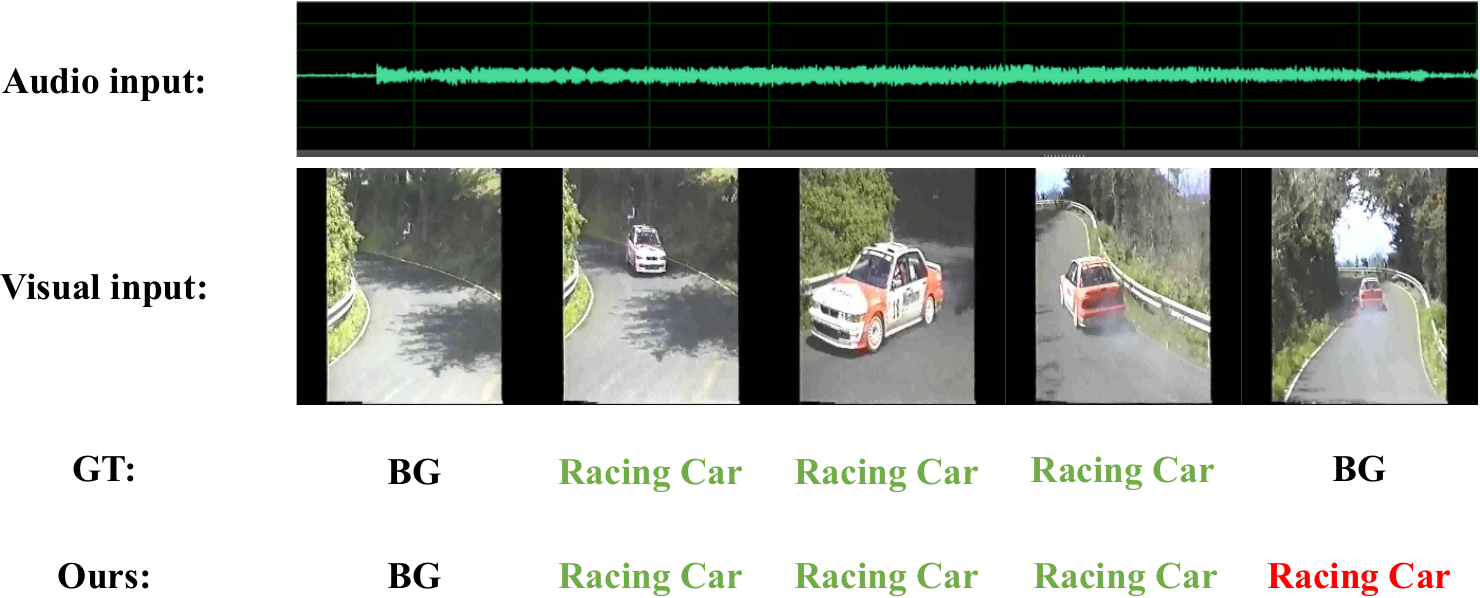}
\renewcommand{\thefigure}{7}
\caption{Illustration of flawed example: ``A racing car approaching and then moving away".}
\label{figc}
\end{figure}

\begin{figure}
\centering
\includegraphics[width=3in]{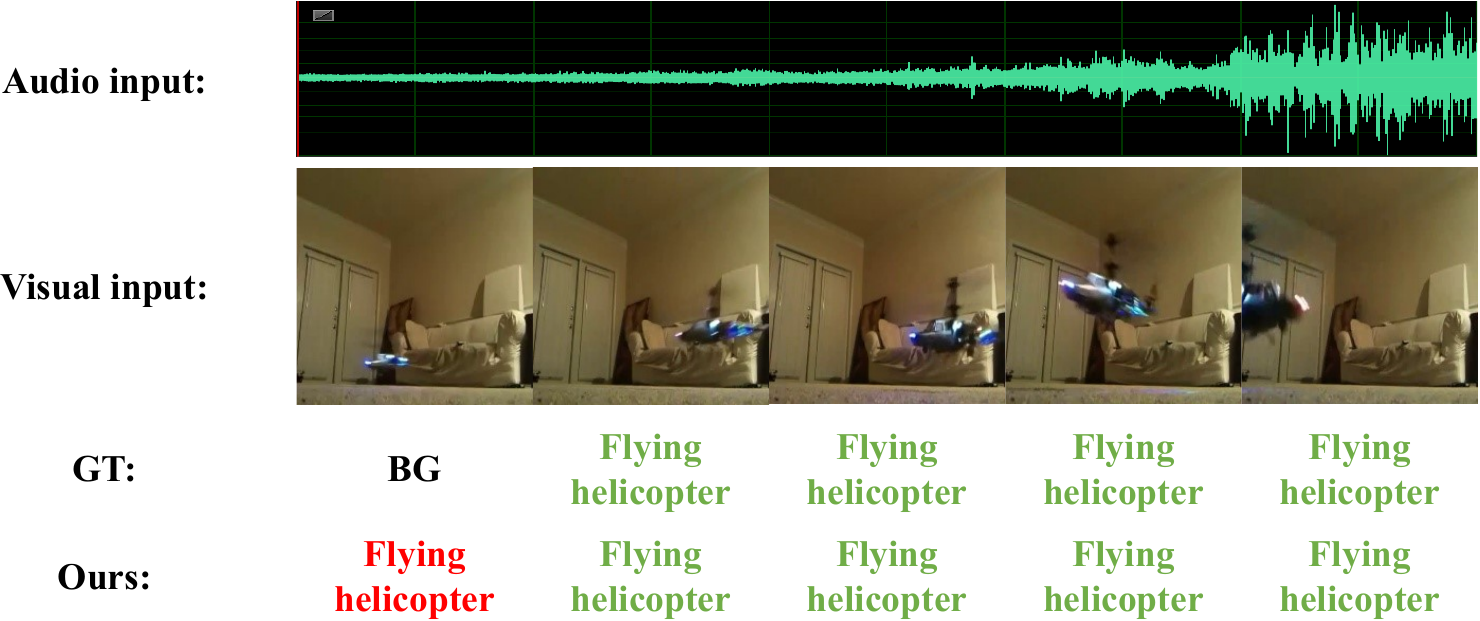}
\renewcommand{\thefigure}{8}
\caption{Illustration of flawed example: ``A helicopter flying in".}
\label{figh}
\end{figure}

VSCG allows the attention module to better focus on the sound-emitting areas of the image rather than only the moving objects (some of which do not make a sound), which are semantically consistent with the video-level event. As shown in Fig. 
\ref{fig4}, for the ``clock" event, the audio contains the continuous ``ticking" sound of the pendulum swinging, the clock striking the hour, and the sound made by the cat when fiddling with the clock and landing. It can be seen that PSP-Net incorrectly locates too much of the sound area on the cat at Seg. 1 when only the ``ticking" sound is present and fails to capture the sound of the cat falling on the ground at Seg. 8. In contrast, AVEL captured the moving object in the picture, i.e., the cat, but incorrectly associated the sound of the clock in the audio with the cat. The 1st and 8th segments vividly demonstrate the convergence and separation of the two sound sources (cat and clock), both of which VSCG has managed to locate more accurately. VSCG cannot only link the primary information contained in the audio, ``the striking", with the real occurring object, ``the clock", but also notice objects which made the sound in the picture due to the video-level guidance.

\subsection{Issues of Temporal Boundaries}
We have found some issues with the labels when exploring the failure examples. As shown in Fig. \ref{figc}, the last segment is tagged with a background category because the visual content, i.e., the car, is too small to recognize. However, our approach successfully identifies that segment as the “racing car” category due to the utilization of video-level semantics. A similar example is shown in Fig. \ref{figh}. The first segment is labeled as a background category even though the helicopter in the scene is both audible and visible. The possible reason for this phenomenon is that the visual target is moving too fast to be recognized by the human eye, so the segment is incorrectly labeled as the background category. However, with our approach, the machine is able to successfully recognize events based on video-level semantics. Exploring more fine-grained temporal boundaries will be an issue to be addressed in future work.

\subsection{Analysis of Failure Results}
\begin{figure}
\centering
\includegraphics[width=3in]{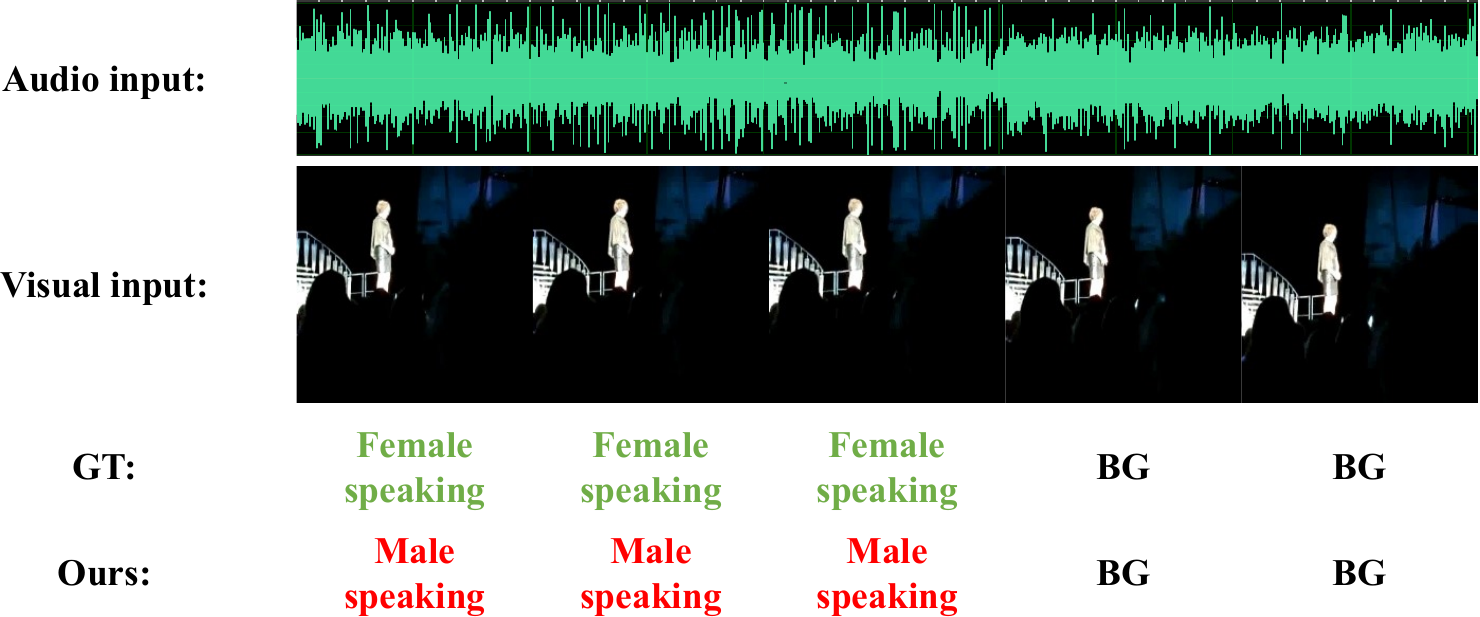}
\renewcommand{\thefigure}{9}
\caption{Illustration of failure example: ``a woman giving a speech on stage, which was filled with applause and cheers".}
\label{figm}
\end{figure}

\begin{figure}
\centering
\includegraphics[width=3in]{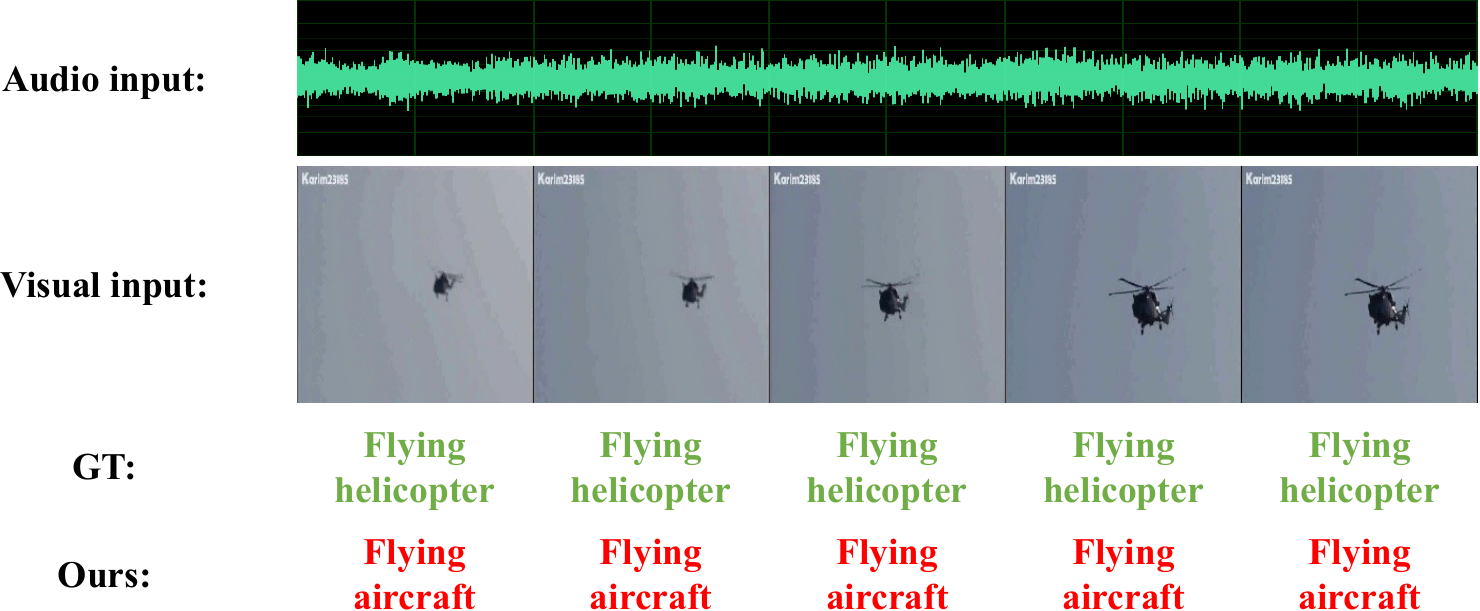}
\renewcommand{\thefigure}{10}
\caption{Illustration of failure example: ``a helicopter hovering overhead in distance".}
\label{figa}
\end{figure}

 Although our proposed semantic consistency-based approach can significantly improve recognition accuracy, it will misidentify events as semantically similar categories when the semantics of the video in both the audio and visual modalities are highly similar to the wrong category. As shown in Fig. \ref{figm} and \ref{figa}, our method incorrectly identifies female speeches as male speeches and the helicopter as aircraft. In the future, the failure rate can be reduced by introducing tags, i.e., text modality that contains explicit semantics, at the training phrase in combination with the semantics we extracted based on the video data.

\end{document}